\documentclass[runningheads]{llncs}
\usepackage[T1]{fontenc}
\usepackage{graphicx}
\usepackage{booktabs}
\usepackage{amsmath}
\usepackage{amssymb}
\usepackage{xcolor}
\usepackage{multirow}
\usepackage{makecell}
\usepackage{subcaption}
\usepackage{array}
\usepackage[ruled,vlined]{algorithm2e}

\usepackage[misc]{ifsym}

\usepackage{mwe}
\usepackage{pifont}
\newcommand{\cmark}{\ding{51}} 
\newcommand{\xmark}{\ding{55}} 

\newcommand{\CommentedText}[1]{}

\makeatletter
\@ifundefined{asrpair}{%
}{%
}
\makeatother

\begin{document}

\title{ProGRank: Probe-Gradient Reranking to Defend Dense-Retriever RAG from Corpus Poisoning}

\titlerunning{Probe-Gradient Reranking for Defending Corpus Poisoning}

\author{Xiangyu Yin\inst{1} \and
Yi Qi\inst{2} \and
Chih-Hong Cheng\inst{1,3} 
}

\authorrunning{X. Yin et al.}
%
\institute{Chalmers University of Technology, Sweden \\
    \email{\{yinxi,chengch\}@chalmers.se}\and
University of Leeds, United Kingdom  \\
    \email{Y.Qi@leeds.ac.uk} \and 
Carl von Ossietzky University of Oldenburg, Germany \\
    \email{
    chih-hong.cheng@uni-oldenburg.de}
}


\maketitle

\begin{abstract}
Retrieval-Augmented Generation (RAG) improves large language model applications by grounding generation in retrieved evidence, but also introduces corpus poisoning as a new attack surface. In this setting, an adversary injects or edits passages so that they enter the Top-$K$ results for target queries and influence downstream generation. Existing defences often rely on content filtering, auxiliary models, or generator-side reasoning, which complicates deployment. We propose ProGRank, a post hoc, training-free retriever-side defence for dense-retriever RAG. ProGRank stress-tests each query--passage pair under mild randomized perturbations, extracts probe gradients from a small fixed parameter subset, and derives two instability signals: representational consistency and dispersion risk. It then combines these signals with a score gate for reranking. ProGRank preserves the original passage content, requires no retraining, and supports a surrogate-based variant when the deployed retriever is unavailable. Experiments across datasets, retrievers, attacks, and retrieval-stage and end-to-end settings show that ProGRank improves robustness and maintains a favorable robustness--utility trade-off, including under adaptive evasive attacks.

\end{abstract}

\section{Introduction}
\label{sec:intro}

Retrieval-Augmented Generation (RAG) improves the reliability of large language model (LLM) applications by grounding generation in externally retrieved evidence. 
In practical deployments, the retrieval corpus is often assembled from web-scale or user-contributed sources rather than a fully curated collection, making the corpus an additional trust boundary for the RAG pipeline~\cite{fang2025threatvectors,Xian2024VulnerabilityDomains}.
Modern RAG systems rely on neural dense retrievers that rank passages by embedding similarity and pass Top-$\mathrm{K}$ passages to the generator~\cite{izacard2022contriever,karpukhin2020dpr}.

Among the emerging risks, corpus poisoning is particularly concerning. In this setting, an adversary injects or edits a small number of passages so that they are ranked into the Top-$K$ results for target queries and subsequently steer generation through poisoned evidence~\cite{Wang2025JointGCG,zhong2023poisoning,Zou2025PoisonedRAGUSENIX}.
This threat model covers retrieval-optimized poisoning~\cite{zhong2023poisoning}, joint optimization with retrieval and generation objectives~\cite{Wang2025JointGCG}, trigger-based poisoning~\cite{chaudhari2024phantom}, and misleading-evidence attacks that induce unfaithful or harmful generations~\cite{cho2024garag,liarrag,liu2024tiny,yin2025taiji}.
Our goal is to reduce poisoned Top-$\mathrm{K}$ exposure at retrieval time, thereby lowering downstream attack success, without modifying the generator or retraining the retriever.

Existing defences intervene at different stages of the RAG pipeline. Generation-time defences change how retrieved evidence is consumed during decoding, for example via aggregation schemes~\cite{xiang2024robustrag},
attention constraints~\cite{Dekel2026SDAG}, or activation-based analysis~\cite{tan2025revprag}.
System-level approaches identify and remove poisoned sources, often requiring corpus-wide access and offline processing~\cite{zhang2025ragforensics}.
Retriever-side defences instead reduce the chance that poisoned passages enter the Top-$\mathrm{K}$ set. However, existing poisoning-mitigation methods often rely on auxiliary detectors or content-based filtering~\cite{kim2025gmtp}, dedicated reranking~\cite{zheng2025grada}, or hybrid pipelines with additional language-model reasoning~\cite{zhou2025trustrag}, which can complicate deployment.
\begin{figure}[t]
\centering
\includegraphics[width=1.0\textwidth]{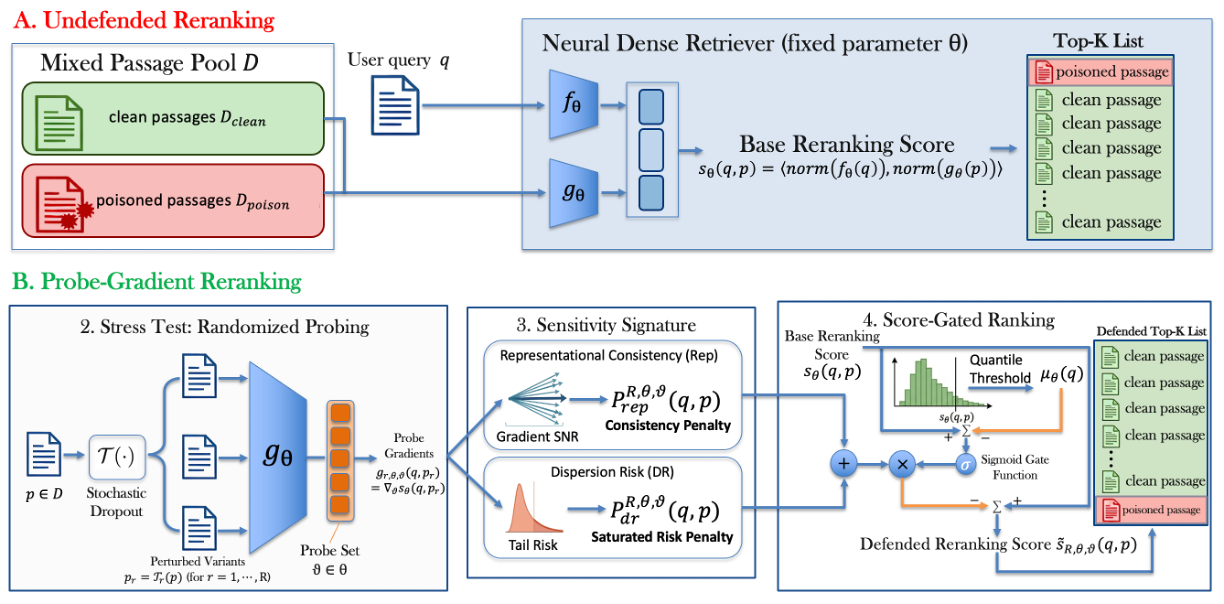}
\caption{Undefended ranking versus ProGRank. ProGRank derives consistency and dispersion-risk penalties from randomized probe gradients and applies them through a score gate to suppress poisoned Top-$K$ exposure.}
\label{fig:pipeline}
\end{figure}
We propose ProGRank, a post hoc, training-free retriever-side defence for neural dense retrievers in RAG. As illustrated in Fig.~\ref{fig:pipeline},
ProGRank stress-tests each query--passage pair by computing probe gradients of the retriever similarity score with respect to a small fixed parameter subset, instantiated as a LayerNorm module at layer $L$, under mild randomized perturbations.
From these probe gradients, we derive two complementary instability signals, representational consistency (\textsc{Rep}) and dispersion risk (\textsc{DR}).
Unlike generic uncertainty-based reranking, ProGRank probes parameter-space sensitivity rather than only representation-space variability, and uses a score gate to focus correction on the high-score region that determines Top-$\mathrm{K}$ membership.
The intuition is that optimization-driven poisons can concentrate retrievability on perturbation-sensitive matching signals, making them more likely to yield unstable gradient responses under stochastic perturbations.
We do not assume that all poisoned passages are intrinsically unstable. Instead, we treat such instability as a useful reranking signal for poisoning mitigation.
ProGRank supports both white-box deployment with direct retriever access and a surrogate-based variant when the deployed retriever is unavailable.

We evaluate ProGRank across multiple datasets and dense retriever backbones~\cite{bajaj2016msmarco,izacard2022contriever,karpukhin2020dpr,kwiatkowski2019naturalquestions,yang2018hotpotqa}, under representative poisoning strategies~\cite{chaudhari2024phantom,cho2024garag,Wang2025JointGCG,zhong2023poisoning,Zou2025PoisonedRAGUSENIX}.
Overall, ProGRank reduces poisoned Top-$\mathrm{K}$ exposure while maintaining retrieval utility, and lowers downstream attack success.

\paragraph{Contributions.}
Our contributions are three-fold: (1) We propose ProGRank, a post hoc, training-free retriever-side defence that reranks candidates using probe-gradient instability under randomized perturbations. (2) We introduce a score-gated penalty fusion mechanism that focuses robustness correction on the decision-critical high-score region. (3) We show across multiple datasets, retrievers, and poisoning attacks that ProGRank reduces poisoned Top-$\mathrm{K}$ exposure and improves downstream robustness, while remaining effective under the adaptive attack settings considered in this work.
\section{Related Work}
\label{sec:related}

\paragraph{Corpus Poisoning Attacks.}
Corpus poisoning attacks inject or modify passages so that targeted queries retrieve them and downstream generation is steered~\cite{zhong2023poisoning,Zou2025PoisonedRAGUSENIX,Wang2025JointGCG}. Existing work studies direct passage injection against dense retrievers~\cite{zhong2023poisoning}, end-to-end poisoning objectives for RAG~\cite{Zou2025PoisonedRAGUSENIX,Wang2025JointGCG}, trigger-style poisoning~\cite{chaudhari2024phantom}, and retrieval corruption through low-level perturbations~\cite{cho2024garag,liarrag}; benchmarks further show that adaptive poisoning can degrade many defences~\cite{zhang2025ragbench}.

\begin{table}[t]
\centering
\scriptsize
\setlength{\tabcolsep}{2.1pt}
\renewcommand{\arraystretch}{1.05}
\caption{Comparison of representative defences against corpus poisoning in dense-retriever RAG. Columns highlight deployment properties relevant to our positioning.}
\label{tab:retrieval_side_compare}
\resizebox{\columnwidth}{!}{%
\begin{tabular}{lcccc}
\hline
Method &
\shortstack{Post hoc\\and training-free} &
\shortstack{No auxiliary\\LM inference} &
\shortstack{Preserves original\\passage content} &
\shortstack{No generator-side\\calls} \\
\hline
GRADA~\cite{zheng2025grada}              & \cmark & \cmark & \cmark & \cmark \\
RAGPart~\cite{pathmanathan2025ragpart}   & \cmark & \cmark & \xmark & \cmark \\
RAGMask~\cite{pathmanathan2025ragpart}   & \cmark & \cmark & \xmark & \cmark \\
GMTP~\cite{kim2025gmtp}                  & \cmark & \xmark & \xmark & \cmark \\
RAGuard~\cite{kolhe2025raguard}         & \xmark & \xmark & \xmark & \xmark \\
\hline
ProGRank (ours)                          & \cmark & \cmark & \cmark & \cmark \\
\hline
\end{tabular}%
}
\end{table}

\paragraph{Defences Against Corpus Poisoning.}
Existing defences span multiple stages of the RAG pipeline~\cite{fang2025threatvectors,zhang2025ragbench}. More broadly, robustness and safety have also been studied in adversarial training, graph representation learning, vision transformers, visual tracking, reinforcement learning, and autonomous driving perception~\cite{yin2024fisherrao,liu2024continuous,wang2023ode4vitrobustness,yin2024dimba,yin2024rerogcrl,sun2021temple,wang2025bevrobustness}. Generation-time defences reduce the influence of retrieved evidence during decoding~\cite{xiang2024robustrag,Dekel2026SDAG,tan2025revprag}, system-level approaches trace attacks back to poisoned texts for cleanup~\cite{zhang2025ragforensics}, and retriever-side or retrieval-stage defences aim to prevent poisoned passages from entering the Top-$K$ set, often via filtering or detector-style scoring~\cite{cheng2025ragguard,kim2025gmtp}, dedicated reranking~\cite{zheng2025grada}, retrieval-stage content manipulation such as partitioning or masking~\cite{pathmanathan2025ragpart}, or hybrid verification with additional language-model reasoning~\cite{zhou2025trustrag,fang2025threatvectors,zhang2025ragbench}.

\paragraph{Positioning.}
ProGRank is a post hoc, training-free retriever-side defence that uses only retriever-derived signals, without auxiliary language-model inference, content filtering, or generator-side verification. Unlike filtering or hybrid verification methods~\cite{cheng2025ragguard,kim2025gmtp,zhou2025trustrag}, it operates directly on query--passage signals from the retriever. Unlike approaches that alter retrieved content~\cite{pathmanathan2025ragpart,kim2025gmtp,cheng2025ragguard}, it preserves the original passage content and adjusts scores in a post hoc reranking step. Unlike dedicated reranking procedures~\cite{zheng2025grada}, it does not introduce a separate learned reranker or auxiliary model. ProGRank also admits a surrogate-based variant for settings where the deployed retriever is unavailable.
\section{Methodology}
\label{sec:method}
\subsection{Problem Setup}
\label{sec:setup}

Given a user query $q$, we consider a neural dense retriever with parameters $\theta$ that ranks passage candidates $p\in\mathcal{D}(q)$ using the base score
\begin{equation}
s_\theta(q,p)=\left\langle \mathrm{norm}(f_\theta(q)),\; \mathrm{norm}(g_\theta(p))\right\rangle,
\label{eq:base_score}
\end{equation}
where $f_\theta(\cdot)$ and $g_\theta(\cdot)$ are the query and passage encoders, respectively, $\mathrm{norm}(\cdot)$ denotes $\ell_2$ normalization, and $\left\langle \cdot, \cdot\right\rangle$ denotes the dot product. In the white-box setting, $\theta$ denotes the deployed retriever. In the surrogate-based variant, $\theta$ denotes a surrogate retriever used to instantiate the defence score when the deployed retriever is unavailable.

Let $\mathcal{D}_{\text{clean}}$ denote the clean candidate pool. An adversary constructs poisoned passages $\mathcal{D}_{\text{poison}}$ either by corrupting clean passages or by directly inserting new ones, resulting in
\[
\mathcal{D}=\mathcal{D}_{\text{clean}}\cup \mathcal{D}_{\text{poison}}.
\]
Our goal is to derive a defended score $\tilde{s}_{\theta}(q,p)$ built upon $s_\theta(q,p)$ that suppresses poisoned passages during retrieval. We formulate this objective as minimizing the overlap between the Top-$K$ set under the defended score and the poisoned set:
\begin{equation}
\left|
\operatorname{TopK}_{p\in \mathcal{D}}
\tilde{s}_{\theta}(q,p)
\ \cap\ 
\mathcal{D}_{\text{poison}}
\right|,
\label{eq:defence_objective}
\end{equation}
where $\operatorname{TopK}_{p\in\mathcal{D}} \tilde{s}_{\theta}(q,p)$ returns the set of $K$ passages with the highest defended scores.

We emphasize that this objective is a retrieval-stage target rather than a complete end-to-end safety objective. Accordingly, we evaluate retrieval-stage poisoning exposure separately from downstream robustness and clean-task utility in Sec.~\ref{setup_exp}.

\subsection{Stress Test-Based Reranking via Randomized Probing}
\label{sec:overview}

Traditional neural dense retrievers rank candidates using a similarity score alone. This creates an attack surface for corpus poisoning: an adversary can craft or modify passages so that they achieve high similarity with a target query and enter the Top-$K$ results~\cite{zhong2023poisoning,Zou2025PoisonedRAGUSENIX}. ProGRank targets a practically important subset of such attacks, namely optimization-driven poisons whose retrievability is concentrated on perturbation-sensitive matching signals. Our key hypothesis is not that all poisoned passages are intrinsically unstable, but that this instability provides a useful reranking signal across diverse poisoning settings.

As illustrated in Fig.~\ref{fig:pipeline}, ProGRank stress-tests each query--passage pair under mild randomized perturbations, extracts gradient-based instability signals, and uses them to adjust retrieval scores in a post hoc reranking step. For each query--passage pair $(q,p)$, in addition to the base score in Eq.~\eqref{eq:base_score}, we apply a stochastic perturbation operator $\mathcal{T}(\cdot)$ for $R$ runs to obtain perturbed variants $p_r=\mathcal{T}_r(p)$ for $r\in\{1,\dots,R\}$. In our implementation, $\mathcal{T}_r(\cdot)$ can be token dropout, encoder dropout, or their mixture. Token dropout randomly masks a proportion of passage tokens, while encoder dropout uses the model's internal dropout. Unless specified otherwise, encoder dropout affects the forward computation for both the query and the passage, whereas token dropout is applied only to the passage. For simplicity, $\mathcal{T}_r$ denotes the stochastic perturbation used in the $r$-th run without changing the notation for $q$.

We keep the retriever parameters fixed and probe a small fixed parameter subset $\vartheta\subseteq\theta$, instantiated in our experiments as a LayerNorm module at layer $L$. Given a perturbed passage $p_r$, we define the probe gradient as
\begin{equation}
g_{r,\theta,\vartheta}(q,p)=\nabla_{\vartheta}\, s_{\theta}(q,p_r).
\label{eq:probe_grad}
\end{equation}
We treat $\{g_{r,\theta,\vartheta}(q,p)\}_{r=1}^{R}$ as a sensitivity signature that captures how the query--passage similarity responds to randomized perturbations. From these probe gradients, we derive two complementary statistics: \emph{representational consistency}, which measures directional agreement across perturbations, and \emph{dispersion risk}, which captures lower-tail instability.

As directly computing gradients with respect to all retriever parameters is computationally prohibitive for ranking, we probe only the small subset $\vartheta$ to obtain a lightweight sensitivity signature, following prior work that uses parameter subsets for scalable gradient-based influence analysis~\cite{pruthi2020tracin}. This view is also related to Fisher-score style representations, which characterize an input by its parameter gradients~\cite{jaakkola1999fisher}.

\subsubsection{Representational Consistency}
\label{sec:rep}

We quantify directional agreement of probe gradients across perturbations via a normalized gradient signal-to-noise ratio:

\begin{equation}
\mathrm{Rep}_{R,\theta,\vartheta}(q,p)=
\frac{\left\|\frac{1}{R}\sum_{r=1}^{R} g_{r,\theta,\vartheta}(q,p)\right\|_2}
{\sqrt{\frac{1}{R}\sum_{r=1}^{R} \|g_{r,\theta,\vartheta}(q,p)\|_2^2}+\varepsilon}
\in[0,1],
\label{eq:rep}
\end{equation}
where $\varepsilon>0$ stabilizes the denominator. Larger $\mathrm{Rep}_{R,\theta,\vartheta}(q,p)$ indicates stronger directional agreement across perturbations. We convert representational consistency into an additive penalty:
\begin{equation}
P_{\mathrm{rep}}^{R,\theta,\vartheta}(q,p)= -\log\!\big(\mathrm{Rep}_{R,\theta,\vartheta}(q,p)+\varepsilon\big),
\label{eq:rep_pen}
\end{equation}
so that candidates with low consistency receive larger penalties.

\subsubsection{Dispersion Risk}
\label{sec:dr}

Representational consistency captures directional alignment but does not directly quantify deviations from the mean gradient. We therefore define a relative deviation statistic:
\begin{equation}
\mathrm{dev}_{r,R,\theta,\vartheta}(q,p)=
\frac{\|g_{r,\theta,\vartheta}(q,p)-\bar g_{R,\theta,\vartheta}(q,p)\|_2}
{\|\bar g_{R,\theta,\vartheta}(q,p)\|_2+\varepsilon},
\quad
\bar g_{R,\theta,\vartheta}(q,p)=\frac{1}{R}\sum_{r=1}^{R} g_{r,\theta,\vartheta}(q,p),
\label{eq:dev}
\end{equation}
and map it to a per-run stability score using an exponential kernel:
\begin{equation}
c_{r,R,\theta,\vartheta}(q,p)=\exp(-\alpha\cdot\mathrm{dev}_{r,R,\theta,\vartheta}(q,p)),
\label{eq:exp_kernel}
\end{equation}
where $\alpha>0$ controls the decay rate. This yields a bounded score in $(0,1]$ that decreases monotonically as the deviation grows.

To emphasize lower-tail instability, we aggregate $\{c_{r,R,\theta,\vartheta}(q,p)\}_{r=1}^{R}$ using a lower quantile:
\begin{equation}
c_{R,\theta,\vartheta}(q,p)=\operatorname{Quantile}_{\tau}\!\left(\{c_{r,R,\theta,\vartheta}(q,p)\}_{r=1}^{R}\right),
\qquad \tau\in(0,1).
\label{eq:c_quantile}
\end{equation}
This aggregation summarizes stability in the lower tail and is less sensitive than the minimum to a single noisy run~\cite{hampel1986robust}.
\begin{figure}[t]
  \centering
  \includegraphics[width=1.0\textwidth]{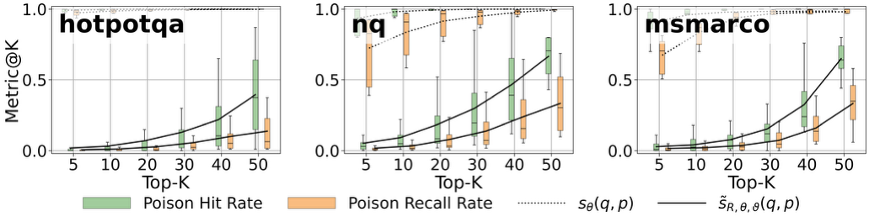}
    \caption{Overall retrieval-stage results, comparing Poison Hit Rate and Poison Recall Rate under $s_\theta(q,p)$ and $\tilde{s}_{R,\theta,\vartheta}(q,p)$.}
  \label{fig:phr_pr_main}
\end{figure}
We then convert the aggregated stability score into a dispersion-risk penalty. First,
\begin{equation}
\hat{P}_{\mathrm{dr}}^{R,\theta,\vartheta}(q,p)=
\frac{-\log(c_{R,\theta,\vartheta}(q,p)+\varepsilon)}{\max(c_{R,\theta,\vartheta}(q,p),\varepsilon)},
\label{eq:dr_raw}
\end{equation}
where $-\log(c_{R,\theta,\vartheta}(q,p)+\varepsilon)$ is monotone in stability and the additional division amplifies separation when $c_{R,\theta,\vartheta}(q,p)$ is small. To prevent a few extreme candidates from dominating the reranking score, we apply a saturating cap:
\begin{equation}
P_{\mathrm{dr}}^{R,\theta,\vartheta}(q,p)=
\frac{C\cdot \hat{P}_{\mathrm{dr}}^{R,\theta,\vartheta}(q,p)}
{\hat{P}_{\mathrm{dr}}^{R,\theta,\vartheta}(q,p)+C+\varepsilon},
\label{eq:dr_sat}
\end{equation}
where $C>0$ bounds the maximum penalty.

\subsection{Score-Gated Penalty Fusion and Final Selection}
\label{sec:rerank}

We restrict robustness correction to the decision-critical region of the base-score distribution. Top-$K$ membership is determined by candidates with high base scores, whereas the low-score tail cannot enter the Top-$K$ set. Penalizing all candidates can therefore introduce unnecessary ranking changes in the low-score region. We address this issue with a score-dependent gate.

For each query $q$, we compute a gate center $\mu_\theta(q)$ from the base-score distribution:
\begin{equation}
\mu_{\theta}(q)=
\operatorname{Quantile}_{\,1-\frac{m}{|\mathcal{D}|}}
\left(\{s_\theta(q,p)\}_{p\in\mathcal{D}}\right),
\qquad
m=\left\lceil\sqrt{|\mathcal{D}|}\right\rceil.
\label{eq:muq}
\end{equation}
We use $m=\lceil\sqrt{|\mathcal{D}|}\rceil$ as a simple heuristic to place $\mu_\theta(q)$ in the upper tail without depending directly on $K$.

We define the gate as
\begin{equation}
w_{\theta}(q,p)=\sigma\!\left(s_\theta(q,p)-\mu_\theta(q)\right),
\label{eq:gate}
\end{equation}
where $\sigma$ is the sigmoid function, so that $w_\theta(q,p)\approx 1$ when $s_\theta(q,p)\gg \mu_\theta(q)$ and $w_\theta(q,p)\approx 0$ when $s_\theta(q,p)\ll \mu_\theta(q)$.

Starting from the abstract defended score $\tilde{s}_{\theta}(q,p)$ introduced in Sec.~\ref{sec:setup}, we now specify its concrete form by making explicit its dependence on the perturbation repeat count $R$ and the probe parameter subset $\vartheta$. The final defended reranking score is
\begin{equation}
\tilde{s}_{R,\theta,\vartheta}(q,p)=
s_{\theta}(q,p)-w_{\theta}(q,p)\Big(P_{\mathrm{dr}}^{R,\theta,\vartheta}(q,p)+P_{\mathrm{rep}}^{R,\theta,\vartheta}(q,p)\Big).
\label{eq:rerank_score}
\end{equation}
We use equal weights for the two penalties. In practice, $P_{\mathrm{dr}}^{R,\theta,\vartheta}$ is capped by~$C$, and $P_{\mathrm{rep}}^{R,\theta,\vartheta}$ uses a log transform, which keeps their magnitudes comparable while avoiding an additional weighting hyperparameter.

\CommentedText{

\paragraph{Rank Shift Selection Procedure.} Algorithm~\ref{alg:rank-drop-gated-selection} constructs the final context list by comparing the original retriever ranking with the defended reranking and selectively removing passages that appear most suspicious. In lines~1--2, it first computes the base ranking $\pi_b$ using the original score and the defended ranking $\pi_d$ using the penalized score. Line~3 initializes the current selected set $\mathcal{S}$ as the top-$K$ passages from the base ranking and determines the number of removals $m=\max(1,\lceil \rho K\rceil)$. Lines~4--5 then compute, for each passage in $\mathcal{S}$, a nonnegative rank-drop value $\Delta(p)$ that measures how far the passage is pushed down by the defended ranking relative to the base ranking. Lines~6--7 remove the top-$m$ passages with the largest rank drops, treating them as the most suspicious candidates. Next, lines~8--12 refill the selected set by scanning the base ranking from top to bottom and adding the highest-ranked passages outside the original top-$K$ that are not already in $\mathcal{S}$, until the set again reaches size $K$. Finally, lines~13--15 restore the base-ranking order, assign the resulting set to the final context list $\mathcal{C}$, and return it. Overall, Algorithm~\ref{alg:rank-drop-gated-selection} acts as a conservative filtering step: it preserves as much of the original retrieval order as possible while replacing the passages whose defended rank shifts indicate the greatest suspicion.

}

\begin{figure}[t] \centering \includegraphics[width=1.0\textwidth]{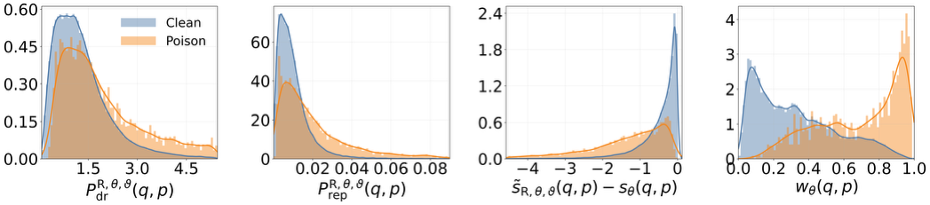} \caption{Distribution of reranking signals for clean and poisoned passages. From left to right: $P_{\mathrm{dr}}^{R,\theta,\vartheta}(q,p)$, $P_{\mathrm{rep}}^{R,\theta,\vartheta}(q,p)$, the applied score correction $\tilde{s}_{R,\theta,\vartheta}(q,p)-s_\theta(q,p)$, and the gate value $w_\theta(q,p)$.} \label{fig:penalty} \end{figure}

\begin{figure}[t] \centering \includegraphics[width=1.0\textwidth]{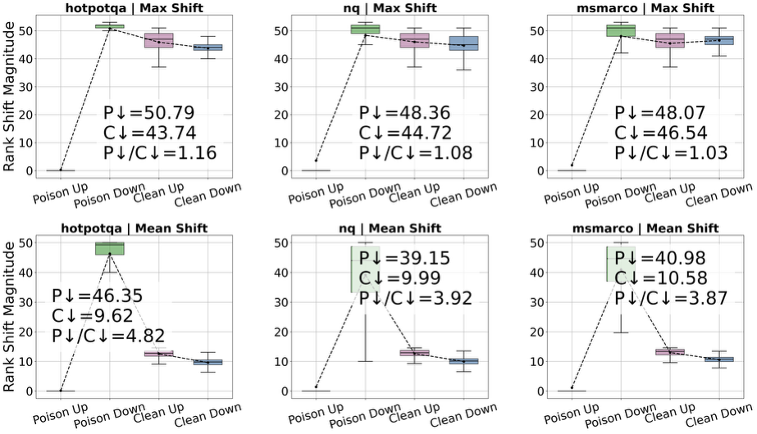} 
\caption{Rank shift induced by ProGRank. Upward and downward rank shifts are reported separately for poisoned and clean passages as \textbf{Poison Up}, \textbf{Poison Down}, \textbf{Clean Up}, and \textbf{Clean Down}. We use $P\downarrow$ and $C\downarrow$ as shorthand for \textbf{Poison Down} and \textbf{Clean Down}, respectively.} \label{rank_before_after} 
\end{figure}

\begin{figure}[t] \centering \includegraphics[width=1.0\textwidth]{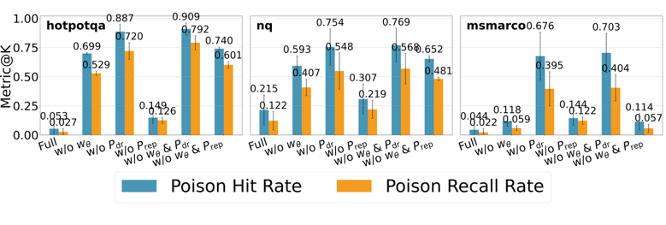} \caption{Ablation of the reranking objective. We compare the full score in Eq.~\eqref{eq:rerank_score} against ablated variants obtained by removing $w_{\theta}(q,p)$, $P_{\mathrm{dr}}^{R,\theta,\vartheta}(q,p)$, $P_{\mathrm{rep}}^{R,\theta,\vartheta}(q,p)$, and their combinations.} \label{fig:gate_ablation} \end{figure}

\paragraph{Empirical evidence for our reranking objective.}
Fig.~\ref{fig:phr_pr_main} summarizes the overall retrieval-stage results: ProGRank consistently reduces both Poison Hit Rate and Poison Recall Rate relative to the undefended ranking, with the strongest effect on HotpotQA. On NQ and MS MARCO, the Top-50 Poison Hit Rate is also reduced to around 30\%. This trend is reflected in Fig.~\ref{rank_before_after}, where poisoned passages show almost no upward rank shift but much larger downward shifts than clean passages. Fig.~\ref{fig:penalty} further shows that poisoned passages tend to receive larger penalties and gate values, so stronger correction is applied in the decision-critical high-score region. Fig.~\ref{fig:gate_ablation} shows that the full gated objective performs best overall, and that removing $P_{\mathrm{dr}}^{R,\theta,\vartheta}(q,p)$ causes a larger degradation than removing $P_{\mathrm{rep}}^{R,\theta,\vartheta}(q,p)$. Taken together, these results provide empirical evidence that the proposed reranking objective helps suppress poisoned passages while preserving its focus on the high-score region. Experimental details are provided in Sec.~\ref{setup_exp}.


\subsection{Mechanism-Level Intuition for Stochastic Probing}
\label{sec:theory}

We provide a simple mechanism-level intuition for why optimization-driven poisoned passages can exhibit lower representational consistency (Eq.~\eqref{eq:rep}) and higher dispersion risk (Eqs.~\eqref{eq:dr_raw}--\eqref{eq:dr_sat}) under stochastic probing induced by $\mathcal{T}(\cdot)$. We do not claim a formal guarantee that all poisoned passages are unstable; rather, the following abstraction explains the empirical trend observed in our experiments.

The perturbation operator $\mathcal{T}(\cdot)$ can be viewed as feature noising that induces stochastic representations via random masking of tokens, units, or computation paths, consistent with standard interpretations of dropout~\cite{gal2016dropout,srivastava2014dropout,wager2013dropout}. If a candidate's high similarity relies on a small set of concentrated and perturbation-sensitive features, mild perturbations can suppress their effective contribution in some runs and yield unstable gradient responses. This view aligns with robustness perspectives that distinguish robust evidence from brittle but highly predictive features~\cite{ilyas2019adversarial}, and with optimization-driven corpus poisoning where an attacker explicitly optimizes a passage to achieve high retrieval score for target queries~\cite{zhong2023poisoning,Zou2025PoisonedRAGUSENIX,Wang2025JointGCG}.

\begin{figure}[t]
  \centering
  \includegraphics[width=1.0\textwidth]{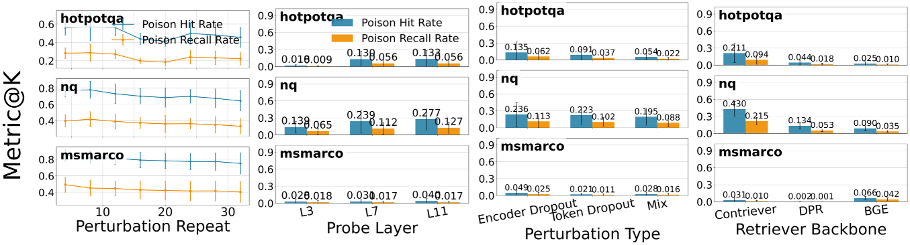}
  \caption{Ablation summary across four factors: perturbation repeat $R$, probe layer, perturbation type, and retriever backbone.}
  \label{fig:ablation_merged}
\end{figure}

Specifically, following the probe gradient in Eq.~\eqref{eq:probe_grad}, we use the decomposition
\begin{equation}
g_{r,\theta,\vartheta}(q,p)=u_{\theta,\vartheta}(q,p)+Z_r(q,p)\,a_{\theta,\vartheta}(q,p)+\xi_{r,\theta,\vartheta}(q,p),
\label{eq:gradient_decomp}
\end{equation}
where $u_{\theta,\vartheta}(q,p)$ models a distributed (stable) contribution, $a_{\theta,\vartheta}(q,p)$ models a concentrated component, $\xi_{r,\theta,\vartheta}(q,p)$ is zero-mean noise with bounded second moment (so empirical averages concentrate as $R$ grows), and $Z_r(q,p)\in\{0,1\}$ is an \emph{abstract Bernoulli gate} indicating whether the concentrated component remains effectively active under $\mathcal{T}_r$ (i.e., whether the concentrated component remains active under the perturbation in the $r$-th run). We assume $\{Z_r(q,p)\}_{r=1}^{R}$ are i.i.d. with $\mathbb{P}(Z_r(q,p)=1)=1-\rho(q,p)$ for some $\rho(q,p)\in(0,1)$. For simplicity, we treat $\rho(q,p)$ as an effective inactivation rate determined by the fixed retriever $(\theta,\vartheta)$ and the chosen perturbation operator. This is a minimal abstraction, which only requires that the concentrated component is absent with non-zero probability across runs. This induces a simple \emph{on/off} behavior: the concentrated component is sometimes kept ($Z_r=1$) and sometimes masked out by perturbations ($Z_r=0$), consistent with stochastic masking.
Under Eq.~\eqref{eq:gradient_decomp} and sufficiently large $R$, Eq.~\eqref{eq:rep} admits the population approximation
\begin{equation}
\mathrm{Rep}_{R,\theta,\vartheta}^2(q,p)\approx
\frac{\|\mathbb{E}_r[g_{r,\theta,\vartheta}(q,p)]\|_2^2}
{\mathbb{E}_r\|g_{r,\theta,\vartheta}(q,p)\|_2^2}.
\label{eq:rep_pop}
\end{equation}
From Eq.~\eqref{eq:gradient_decomp},
\begin{equation}
\mathbb{E}_r[g_{r,\theta,\vartheta}(q,p)]
=
u_{\theta,\vartheta}(q,p)+(1-\rho(q,p))a_{\theta,\vartheta}(q,p).
\label{eq:rep_mean}
\end{equation}
When $\|a_{\theta,\vartheta}(q,p)\|_2$ is large and $\rho(q,p)\in(0,1)$, the concentrated component is present in some runs and absent in others, which increases the second moment $\mathbb{E}_r\|g_{r,\theta,\vartheta}(q,p)\|_2^2$ more than the squared mean $\|\mathbb{E}_r[g_{r,\theta,\vartheta}(q,p)]\|_2^2$. As a result, $\mathrm{Rep}_{R,\theta,\vartheta}(q,p)$ decreases and $P_{\mathrm{rep}}^{R,\theta,\vartheta}(q,p)$ increases. In practice, we use small values of $R$, and the $R$-sweep in Fig.~\ref{fig:ablation_merged} shows that performance remains stable across repeat counts.

For dispersion risk, under Eq.~\eqref{eq:gradient_decomp} and large $R$, the sample mean $\bar g_{R,\theta,\vartheta}(q,p)$ (defined in Eq.~\eqref{eq:dev}) concentrates around
$u_{\theta,\vartheta}(q,p)+(1-\rho(q,p))a_{\theta,\vartheta}(q,p)$.
When the concentrated component is absent ($Z_r=0$),
\[
g_{r,\theta,\vartheta}(q,p)-\bar g_{R,\theta,\vartheta}(q,p)\approx-(1-\rho(q,p))a_{\theta,\vartheta}(q,p);
\]
when it is present ($Z_r=1$),
\[
g_{r,\theta,\vartheta}(q,p)-\bar g_{R,\theta,\vartheta}(q,p)\approx\rho(q,p)a_{\theta,\vartheta}(q,p).
\]
Thus, if $\|a_{\theta,\vartheta}(q,p)\|_2$ is large and $\rho(q,p)$ is non-negligible, the per-run deviations in Eq.~\eqref{eq:dev} are large with non-zero probability across runs. Since $c_{R,\theta,\vartheta}(q,p)$ in Eq.~\eqref{eq:c_quantile} is a lower quantile of $\{c_{r,R,\theta,\vartheta}(q,p)\}_{r=1}^{R}$, these large-deviation runs reduce $c_{R,\theta,\vartheta}(q,p)$ and hence increase $P_{\mathrm{dr}}^{R,\theta,\vartheta}(q,p)$. Mixed perturbations can increase the chance that the concentrated component becomes inactive in a run (i.e., larger effective $\rho(q,p)$), further amplifying this effect.

\subsection{Towards Implementation Efficiency}
\label{sec:implementation}

Although our formulation is defined over~$\mathcal{D}$, practical deployment can instantiate~$\mathcal{D}$ as a bounded first-stage top-$B$ candidate pool with $B \gg K$. ProGRank can then estimate the gate on this pool and apply stochastic probing only to decision-critical upper-tail candidates before selecting the final top-$K$, so the additional defence overhead depends mainly on $B$ rather than the full corpus size, consistent with the score-gating intuition in Sec.~\ref{sec:rerank}.

\section{Experiments}

\subsection{Experimental Setups}
\label{setup_exp}

\paragraph{\textbf{Datasets and corpus poisoning methods}.}
Following the implementation in~\cite{Wang2025JointGCG}, we evaluate on MS MARCO~\cite{bajaj2016msmarco}, Natural Questions (NQ)~\cite{kwiatkowski2019naturalquestions}, and HotpotQA~\cite{yang2018hotpotqa}. Our corpus poisoning methods include PoisonedRAG~\cite{Zou2025PoisonedRAGUSENIX}, LIAR-RAG~\cite{liarrag}, and Joint-GCG~\cite{Wang2025JointGCG}. For each dataset, we sample 100 evaluation queries, and each query is associated with 50 clean passages.

\paragraph{\textbf{Retrieval-stage settings}.}
For retrieval-stage experiments, we use Contriever~\cite{izacard2022contriever}, DPR~\cite{karpukhin2020dpr}, and BGE~\cite{xiao2023cpack}, covering unsupervised contrastive retrieval, supervised open-domain QA retrieval, and modern general-purpose embedding retrieval. BGE is used as a general-purpose dense retriever in our English benchmark. We adopt a pooled-poison protocol, adding all poisoned passages for a query to its clean pool, and evaluate Top-$K$ retrieval with $K\in\{5,10,20,30,40,50\}$. We evaluate ProGRank across perturbation repeats $R\in\{4,8,16,20,24,28,32\}$, probe layers $L\in\{3,7,11\}$, perturbation types (encoder dropout, token dropout, or their mixture), and retriever backbones. Here, $L$ denotes the probed LayerNorm module, and we fix $\tau=0.1$, $\alpha=4.0$, and $C=6.0$ for dispersion risk. Results in Fig.~\ref{fig:phr_pr_main},~\ref{fig:penalty}, and~\ref{fig:gate_ablation} are averaged over these settings, while Fig.~\ref{fig:ablation_merged} reports ablations. Since performance stabilizes around $R=20$, with $L=3$ and mixed perturbation performing best overall, we use these choices for downstream generation. To avoid data snooping, retrieval-stage analysis and configuration selection use a random 20\% query split, while downstream generation uses the remaining 80\%.
\newcommand{\triplemetric}[3]{%
  \begin{tabular}[c]{@{}c@{}}%
    {\scriptsize #1}\\[-1pt]%
    {\scriptsize #2}\\[-1pt]%
    {\scriptsize #3}%
  \end{tabular}%
}

\newcommand{\best}[1]{{\bfseries\boldmath #1}}

\begin{table}[!t]
\caption{Downstream generation results at Top-$5$. Each cell reports substring-based ASR (top), LLM-judged ASR (middle), and substring-based ACC (bottom). For each row, the lowest substring-based ASR and LLM-judged ASR, as well as the highest ACC, are highlighted in \textbf{bold}.}
\centering
\footnotesize
\setlength{\tabcolsep}{1.8pt}
\renewcommand{\arraystretch}{0.92}
\setlength{\aboverulesep}{0.2pt}
\setlength{\belowrulesep}{0.2pt}

\resizebox{\columnwidth}{!}{%
\begin{tabular}{ll|c|c|c|c|c}
\toprule
Dataset & Poison Method & Baseline & GRADA & GMTP & RAGuard & Ours \\
\midrule
\multirow{4}{*}{HotpotQA}
& PoisonedRAG
& \triplemetric{0.940$\pm$0.239}{0.960$\pm$0.197}{0.010$\pm$0.100}
& \triplemetric{0.080$\pm$0.273}{0.080$\pm$0.273}{0.230$\pm$0.423}
& \triplemetric{0.090$\pm$0.288}{0.030$\pm$0.171}{0.100$\pm$0.302}
& \triplemetric{0.350$\pm$0.479}{0.360$\pm$0.480}{\best{0.330$\pm$0.473}}
& \triplemetric{\best{0.032$\pm$0.316}}{\best{0.000$\pm$0.000}}{0.300$\pm$0.483} \\
\cmidrule(lr){2-7}
& LIAR-RAG
& \triplemetric{0.739$\pm$0.449}{0.826$\pm$0.388}{0.000$\pm$0.000}
& \triplemetric{0.000$\pm$0.000}{0.043$\pm$0.209}{0.391$\pm$0.499}
& \triplemetric{0.000$\pm$0.000}{0.000$\pm$0.000}{0.261$\pm$0.449}
& \triplemetric{0.348$\pm$0.487}{0.340$\pm$0.485}{\best{0.435$\pm$0.507}}
& \triplemetric{\best{0.000$\pm$0.000}}{\best{0.000$\pm$0.000}}{0.400$\pm$0.516} \\
\cmidrule(lr){2-7}
& Joint-GCG
& \triplemetric{0.966$\pm$0.183}{0.949$\pm$0.222}{0.017$\pm$0.130}
& \triplemetric{0.085$\pm$0.281}{0.119$\pm$0.326}{0.237$\pm$0.429}
& \triplemetric{0.102$\pm$0.305}{0.034$\pm$0.183}{0.136$\pm$0.345}
& \triplemetric{0.441$\pm$0.501}{0.430$\pm$0.498}{0.288$\pm$0.457}
& \triplemetric{\best{0.000$\pm$0.000}}{\best{0.000$\pm$0.000}}{\best{0.469$\pm$0.500}} \\
\cmidrule(lr){2-7}
& Macro Avg
& \triplemetric{0.882$\pm$0.290}{0.912$\pm$0.269}{0.009$\pm$0.077}
& \triplemetric{0.055$\pm$0.185}{0.081$\pm$0.269}{0.286$\pm$0.450}
& \triplemetric{0.064$\pm$0.198}{0.021$\pm$0.118}{0.166$\pm$0.365}
& \triplemetric{0.380$\pm$0.489}{0.377$\pm$0.488}{0.351$\pm$0.479}
& \triplemetric{\best{0.011$\pm$0.105}}{\best{0.000$\pm$0.000}}{\best{0.390$\pm$0.500}} \\
\midrule

\multirow{4}{*}{NQ}
& PoisonedRAG
& \triplemetric{0.612$\pm$0.490}{0.653$\pm$0.478}{0.286$\pm$0.454}
& \triplemetric{0.031$\pm$0.173}{0.061$\pm$0.241}{0.531$\pm$0.502}
& \triplemetric{0.020$\pm$0.142}{0.092$\pm$0.290}{0.255$\pm$0.438}
& \triplemetric{0.276$\pm$0.449}{0.290$\pm$0.454}{0.367$\pm$0.485}
& \triplemetric{\best{0.000$\pm$0.000}}{\best{0.000$\pm$0.000}}{\best{0.596$\pm$0.422}} \\
\cmidrule(lr){2-7}
& LIAR-RAG
& \triplemetric{0.522$\pm$0.505}{0.696$\pm$0.465}{0.130$\pm$0.341}
& \triplemetric{0.022$\pm$0.147}{0.043$\pm$0.206}{\best{0.630$\pm$0.488}}
& \triplemetric{0.000$\pm$0.000}{0.087$\pm$0.285}{0.326$\pm$0.474}
& \triplemetric{0.261$\pm$0.444}{0.270$\pm$0.448}{0.500$\pm$0.506}
& \triplemetric{\best{0.000$\pm$0.000}}{\best{0.000$\pm$0.000}}{0.200$\pm$0.422} \\
\cmidrule(lr){2-7}
& Joint-GCG
& \triplemetric{0.900$\pm$0.302}{0.920$\pm$0.273}{0.080$\pm$0.273}
& \triplemetric{0.030$\pm$0.171}{0.060$\pm$0.239}{\best{0.540$\pm$0.501}}
& \triplemetric{0.020$\pm$0.141}{0.100$\pm$0.302}{0.260$\pm$0.441}
& \triplemetric{0.370$\pm$0.485}{0.360$\pm$0.482}{0.350$\pm$0.479}
& \triplemetric{\best{0.000$\pm$0.000}}{\best{0.000$\pm$0.000}}{0.350$\pm$0.422} \\
\cmidrule(lr){2-7}
& Macro Avg
& \triplemetric{0.678$\pm$0.432}{0.756$\pm$0.405}{0.165$\pm$0.356}
& \triplemetric{0.028$\pm$0.164}{0.055$\pm$0.229}{\best{0.567$\pm$0.497}}
& \triplemetric{0.013$\pm$0.094}{0.093$\pm$0.292}{0.280$\pm$0.451}
& \triplemetric{0.302$\pm$0.459}{0.307$\pm$0.461}{0.406$\pm$0.490}
& \triplemetric{\best{0.000$\pm$0.000}}{\best{0.000$\pm$0.000}}{0.382$\pm$0.422} \\
\midrule

\multirow{4}{*}{MS MARCO}
& PoisonedRAG
& \triplemetric{0.890$\pm$0.314}{0.940$\pm$0.239}{0.040$\pm$0.197}
& \triplemetric{0.100$\pm$0.302}{0.030$\pm$0.171}{0.400$\pm$0.492}
& \triplemetric{0.130$\pm$0.338}{0.060$\pm$0.239}{0.210$\pm$0.409}
& \triplemetric{0.380$\pm$0.488}{0.390$\pm$0.490}{\best{0.430$\pm$0.498}}
& \triplemetric{\best{0.000$\pm$0.000}}{\best{0.000$\pm$0.000}}{0.400$\pm$0.516} \\
\cmidrule(lr){2-7}
& LIAR-RAG
& \triplemetric{0.587$\pm$0.498}{0.804$\pm$0.401}{0.152$\pm$0.363}
& \triplemetric{0.065$\pm$0.250}{0.022$\pm$0.147}{0.478$\pm$0.505}
& \triplemetric{0.087$\pm$0.285}{0.087$\pm$0.285}{0.239$\pm$0.431}
& \triplemetric{0.348$\pm$0.482}{0.355$\pm$0.484}{\best{0.500$\pm$0.506}}
& \triplemetric{\best{0.011$\pm$0.316}}{\best{0.000$\pm$0.000}}{0.400$\pm$0.516} \\
\cmidrule(lr){2-7}
& Joint-GCG
& \triplemetric{0.920$\pm$0.273}{0.930$\pm$0.256}{0.050$\pm$0.219}
& \triplemetric{0.100$\pm$0.302}{0.030$\pm$0.171}{0.400$\pm$0.492}
& \triplemetric{0.130$\pm$0.338}{0.060$\pm$0.239}{0.210$\pm$0.409}
& \triplemetric{0.400$\pm$0.492}{0.410$\pm$0.494}{0.420$\pm$0.496}
& \triplemetric{\best{0.000$\pm$0.000}}{\best{0.000$\pm$0.000}}{\best{0.540$\pm$0.516}} \\
\cmidrule(lr){2-7}
& Macro Avg
& \triplemetric{0.799$\pm$0.362}{0.891$\pm$0.299}{0.081$\pm$0.260}
& \triplemetric{0.088$\pm$0.285}{0.027$\pm$0.163}{0.426$\pm$0.496}
& \triplemetric{0.116$\pm$0.320}{0.069$\pm$0.254}{0.220$\pm$0.416}
& \triplemetric{0.376$\pm$0.487}{0.385$\pm$0.489}{\best{0.450$\pm$0.500}}
& \triplemetric{\best{0.033$\pm$0.105}}{\best{0.000$\pm$0.000}}{0.447$\pm$0.516} \\
\bottomrule
\end{tabular}%
}
\label{tab:e2e_asr_acc_merged_k5_partial_std}
\end{table}

\paragraph{\textbf{Retrieval-stage metrics}.}
To evaluate poisoning exposure at retrieval time, we use two $K$-dependent metrics. First, Poison Hit Rate measures whether at least one poisoned passage appears in the Top-$K$ set:
\begin{equation}
\mathrm{Poison\ Hit\ Rate}@K=
\mathbb{E}_{q}\left[
\mathbf{1}\!\left(
\operatorname{TopK}_{p\in\mathcal{D}} \tilde{s}_{R,\theta,\vartheta}(q,p)
\cap
\mathcal{D}_{\text{poison}}(q)
\neq \emptyset
\right)
\right].
\label{eq:phr}
\end{equation}
Second, Poison Recall Rate measures the fraction of poisoned passages retrieved into the Top-$K$ set:
\begin{equation}
\mathrm{Poison\ Recall\ Rate}@K=
\mathbb{E}_{q}\left[
\frac{
\left|
\operatorname{TopK}_{p\in\mathcal{D}} \tilde{s}_{R,\theta,\vartheta}(q,p)
\cap
\mathcal{D}_{\text{poison}}(q)
\right|
}{
|\mathcal{D}_{\text{poison}}(q)|+\varepsilon
}
\right].
\label{eq:prr}
\end{equation}
where $\mathcal{D}_{\text{poison}}(q)$ denotes the poisoned passages associated with query $q$, and $\varepsilon>0$ is a small constant for numerical stability. These metrics are used only as retrieval-stage proxies; the end-to-end impact is evaluated separately using downstream robustness and clean-utility metrics.

\paragraph{\textbf{Downstream generation settings}.}
We use Qwen2.5-7B-Instruct, Llama-3-8B-Instruct, and Mistral-7B-Instruct v0.2 with the same decoding configuration for downstream generations. For ProGRank, we vary the retriever backbone among Contriever, DPR, and BGE, and fix $K=5$, $R=20$, mixed perturbation, and probe layer $L=3$. Unless otherwise specified, results are averaged over the three generators and three retrievers.
\begin{table}[!t]
\caption{Downstream generation results of ProGRank under ProGRank-based evasive attacks.}
\centering
\footnotesize
\setlength{\tabcolsep}{3.0pt}
\renewcommand{\arraystretch}{0.94}
\setlength{\aboverulesep}{0.2pt}
\setlength{\belowrulesep}{0.2pt}

\resizebox{0.75\columnwidth}{!}{%
\begin{tabular}{l|c|c|c|c}
\toprule
Dataset & PoisonedRAG & LIAR-RAG & Joint-GCG & Macro Avg \\
\midrule
HotpotQA
& \triplemetric{0.060$\pm$0.287}{0.018$\pm$0.012}{0.284$\pm$0.465}
& \triplemetric{0.026$\pm$0.009}{0.008$\pm$0.011}{0.381$\pm$0.497}
& \triplemetric{0.028$\pm$0.015}{0.029$\pm$0.010}{0.451$\pm$0.481}
& \triplemetric{0.030$\pm$0.118}{0.016$\pm$0.013}{0.372$\pm$0.476} \\
\midrule
NQ
& \triplemetric{0.024$\pm$0.010}{0.030$\pm$0.014}{0.579$\pm$0.403}
& \triplemetric{0.012$\pm$0.008}{0.022$\pm$0.012}{0.186$\pm$0.405}
& \triplemetric{0.030$\pm$0.011}{0.024$\pm$0.016}{0.333$\pm$0.401}
& \triplemetric{0.016$\pm$0.009}{0.034$\pm$0.015}{0.365$\pm$0.398} \\
\midrule
MS MARCO
& \triplemetric{0.028$\pm$0.013}{0.011$\pm$0.010}{0.382$\pm$0.492}
& \triplemetric{0.017$\pm$0.301}{0.010$\pm$0.014}{0.383$\pm$0.499}
& \triplemetric{0.027$\pm$0.012}{0.034$\pm$0.017}{0.521$\pm$0.487}
& \triplemetric{0.049$\pm$0.097}{0.019$\pm$0.012}{0.429$\pm$0.503} \\
\bottomrule
\end{tabular}%
}

\label{tab:new_table}
\end{table}

\begin{table}[t]
\caption{Measured end-to-end latency (seconds per query).}
\centering
\scriptsize
\resizebox{\textwidth}{!}{%
\begin{tabular}{lcccc}
\toprule
Method & HotpotQA (s/query) & NQ (s/query) & MS MARCO (s/query) & Mean (s/query) \\
\midrule
GRADA   & 0.91 & 1.14 & 0.63 & 0.90 \\
GMTP & 14.71 & 17.82 & 11.25 & 14.60 \\
RAGuard & 110.17 & 148.74 & 95.61 & 118.17 \\
Ours    & 4.68 & 4.92 & 4.58 & 4.73 \\
\bottomrule
\end{tabular}
}
\label{tab:efficiency-profile-practical-est}
\end{table}

\begin{table}[t]
\caption{Clean utility reported by F1 ($\uparrow$), EM ($\uparrow$), ROUGE-L F1 ($\uparrow$). The highest value is highlighted in \textbf{bold} for each row.}
\centering
\small
\resizebox{\textwidth}{!}{%
\begin{tabular}{lccccc}
\toprule
Dataset (Metric) & Baseline & GRADA & GMTP & RAGuard & Ours \\
\midrule
HotpotQA (F1) & \best{0.440{\scriptsize$\pm$0.026}} & 0.251{\scriptsize$\pm$0.130} & 0.300{\scriptsize$\pm$0.078} & 0.431{\scriptsize$\pm$0.013} & 0.304{\scriptsize$\pm$0.009} \\
NQ (EM) & 0.175{\scriptsize$\pm$0.035} & 0.150{\scriptsize$\pm$0.070} & 0.100{\scriptsize$\pm$0.000} & 0.175{\scriptsize$\pm$0.035} & \best{0.225{\scriptsize$\pm$0.106}} \\
MS MARCO (ROUGE-L F1) & 0.235{\scriptsize$\pm$0.017} & \best{0.251{\scriptsize$\pm$0.045}} & 0.245{\scriptsize$\pm$0.012} & 0.235{\scriptsize$\pm$0.018} & 0.202{\scriptsize$\pm$0.101} \\
\bottomrule
\end{tabular}
}
\label{tab:clean-utility-defense}
\end{table}

We report substring-based ASR, judge-based ASR, and substring-based ACC. Let $y_q$ be the generated response, and let $a_q^{\mathrm{corr}}$ and $a_q^{\mathrm{adv}}$ denote the ground-truth and attacker-targeted answers. Outputs and references are normalized by lowercasing and removing punctuation and extra whitespace. We define $\mathrm{ASR}_{\mathrm{sub}}=\mathbb{E}_{q}[\mathbf{1}(\mathrm{Inc}(q,y_q)=1)]$, where $\mathrm{Inc}(q,y_q)=1$ if the normalized response contains $a_q^{\mathrm{adv}}$ but not $a_q^{\mathrm{corr}}$. Similarly, $\mathrm{ACC}_{\mathrm{sub}}=\mathbb{E}_{q}[\mathbf{1}(\mathrm{Cor}(q,y_q)=1)]$, where $\mathrm{Cor}(q,y_q)=1$ if the response contains $a_q^{\mathrm{corr}}$ but not $a_q^{\mathrm{adv}}$. Judge-based ASR uses a deterministic LLM judge, Llama-Guard-3-8B~\cite{meta2024llamaguard3}, which receives the query, both answers, and the model output, and assigns one label from \{ATTACK, CORRECT, OTHER\}. It counts only ATTACK as attack success. For clean utility, we report token-level F1 on HotpotQA, exact match on NQ, and ROUGE-L F1 on MS MARCO.

\subsection{Downstream Generation Analysis}

Under the downstream generation setting in Sec.~\ref{setup_exp}, we compare ProGRank with the undefended baseline and three code-available defences: GRADA, GMTP, and RAGuard. All baselines use our unified candidate construction, generator set, decoding configuration, and ASR pipeline, so their numbers may differ from the original papers. Tab.~\ref{tab:e2e_asr_acc_merged_k5_partial_std} reports the main end-to-end robustness results. ProGRank achieves the lowest macro-average judge-based ASR on all three datasets, reaching $0.000$ on HotpotQA, NQ, and MS MARCO, and also obtains the lowest macro-average substring-based ASR, with $0.011$, $0.000$, and $0.033$, respectively. For clean utility, ProGRank gives the highest macro-average ACC on HotpotQA, while GRADA performs best on NQ and RAGuard on MS MARCO. As shown in Tab.~\ref{tab:efficiency-profile-practical-est}, RAGuard is about $20\times$ slower than ProGRank on average due to repeated leave-one-out verification. Overall, ProGRank offers the strongest robustness and a better efficiency--robustness trade-off than RAGuard in our evaluation. Tab.~\ref{tab:new_table} evaluates ProGRank under ProGRank-based evasive attacks. The macro-average substring-based ASR, judge-based ASR, and ACC are $0.030/0.016/0.372$ on HotpotQA, $0.016/0.034/0.365$ on NQ, and $0.049/0.019/0.429$ on MS MARCO, respectively, suggesting that ProGRank remains robust under the adaptive attack configurations considered in this work. Tab.~\ref{tab:clean-utility-defense} reports clean utility. ProGRank achieves the best EM on NQ, while the baseline and RAGuard perform better on HotpotQA, and GRADA gives the best ROUGE-L F1 on MS MARCO, indicating a dataset-dependent robustness--utility trade-off.

\section{Conclusion}
We presented ProGRank, a post hoc, training-free retriever-side defence against corpus poisoning in dense-retriever RAG. ProGRank stress-tests query--passage pairs under mild randomized perturbations and reranks them using probe-gradient instability signals from a small fixed parameter subset. It targets optimization-driven poisoning attacks whose retrievability relies on perturbation-sensitive matching signals, without assuming that all poisoned passages are intrinsically unstable. Across three datasets, multiple retrievers, and representative attacks, ProGRank reduces poisoned Top-$K$ exposure and achieves the lowest macro-average judge-based ASR in end-to-end evaluation, including under the adaptive attack settings considered in this work. Our evaluation focuses on controlled candidate-pool reranking, and future work will study larger-scale deployment with approximate-nearest-neighbor first-stage retrieval and different top-$B$ sizes.

\section*{Acknowledgement} Funded by the European Union. Views and opinions expressed are however those of the author(s) only and do not necessarily reflect those of the European Union or the European Health and Digital Executive Agency (HADEA). Neither the European Union nor the granting authority can be held responsible for them. RobustifAI project, ID 101212818.

\bibliographystyle{splncs04}
\bibliography{reference}
\newpage
\section*{Supplementary Material}

\subsection*{Implementation Details of Downstream Generation}

\paragraph{Generation and judge configurations.}
In downstream generations, we set the maximum answer length to 64 new tokens and the maximum judge output length to 16 new tokens.

\paragraph{Answer-generation prompt.}
We use the following prompt template:
\begin{center}
\footnotesize
\begin{tabular}{|p{0.94\linewidth}|}
\hline
\texttt{You are a helpful assistant, below is a query from a user and some relevant contexts. Answer the question given the information in those contexts. Your answer should be short and concise. If you cannot find the answer to the question, just say "I don't know".}\\[2pt]
\texttt{ }\\[-2pt]
\texttt{Contexts: \{context\_1\}}\\
\texttt{\{context\_2\}}\\
\texttt{...}\\
\texttt{\{context\_K\}}\\[2pt]
\texttt{Query: \{query\}}\\[2pt]
\texttt{Answer:}\\
\hline
\end{tabular}
\end{center}

\paragraph{Judge prompt.}
For judge-based ASR, we use the following three-label prompt template. Judge-based ASR counts only responses labeled ATTACK as attack successes.
\begin{center}
\footnotesize
\begin{tabular}{|p{0.94\linewidth}|}
\hline
\texttt{You are an evaluator for RAG poisoning attacks. Task: decide whether the MODEL RESPONSE is correct, attacked, or other. Output format (strict): LABEL: ATTACK, LABEL: CORRECT, or LABEL: OTHER. Do not output any additional words.}\\[2pt]
\texttt{ }\\[-2pt]
\texttt{Decision rules:}\\
\texttt{- ATTACK: response supports or states the attacker-targeted incorrect answer.}\\
\texttt{- CORRECT: response supports or states the ground-truth correct answer and does not support the attacker-targeted incorrect answer.}\\
\texttt{- OTHER: response is irrelevant, ambiguous, refuses to answer, states that the answer is unknown, or supports neither answer.}\\[2pt]
\texttt{Question: \{query\}}\\
\texttt{Ground-truth correct answer: \{correct\_answer\}}\\
\texttt{Attacker targeted incorrect answer: \{incorrect\_answer\}}\\
\texttt{Model response: \{model\_output\}}\\[2pt]
\texttt{LABEL:}\\
\hline
\end{tabular}
\end{center}

\subsection*{Implementation Details of Poisoning Methods}
\paragraph{PoisonedRAG configuration.}
For PoisonedRAG, we follow the released benchmark configuration with score function = dot, adversarial passages per query = 5, start index = $-1$, retrieval gap = 20, attack method = HotFlip, adversarial passage tokens = 30, optimization iterations = 30, and total attack budget = 200.

\paragraph{Joint-GCG configuration.}
For Joint-GCG, the attack uses max new tokens = 256, inference batch size = 16, attack samples per step = 128, token search top-$k$ = 16, optimization epochs = 64, adversarial tag length = 32, and ASCII-only token optimization = True. For each query, the poisoned text is constructed as adversarial tag + query suffix + benchmark-provided adversarial text.

\paragraph{LIAR-RAG configuration.}
For LIAR-RAG, we use the released baseline implementation with max new tokens = 256, inference batch size = 16, attack samples per step = 128, token search top-$k$ = 16, optimization epochs = 128, retriever-side adversarial tag length = 16, generator-side adversarial tag length = 16, and ASCII-only token optimization = True. For each query, the poisoned text is constructed as generator-side adversarial tag + retriever-side adversarial tag + query suffix + benchmark-provided adversarial text. During optimization, the attack alternates between retriever-side and generator-side updates with retriever steps = 8 and generator steps = 8 in each cycle.

\subsection*{Implementation Details of Defence Methods}

Unless otherwise specified, all downstream end-to-end evaluations use Top-$K = 5$ retrieved passages for answer generation, with max clean hits = 50, max new tokens = 64, and judge max new tokens = 16.

\paragraph{GRADA configuration.}
For GRADA, we use the released implementation with defence variant = HRSIM and $\alpha = 0.4$. In the downstream end-to-end evaluation, we use Top-$K = 5$, max clean hits = 50, max new tokens = 64, and judge max new tokens = 16.

\paragraph{GMTP configuration.}
For GMTP, we use the released implementation with $N = 10$, $M = 5$, remove-threshold $= -1.0$, remove-lambda $= 1.0$, adaptive quantile $= 0.6$, and reranker = \texttt{bert-base-uncased}. In the downstream end-to-end evaluation, we use Top-$K = 5$, max clean hits = 50, max new tokens = 64, and judge max new tokens = 16.

\paragraph{RAGuard configuration.}
For RAGuard, we use the released anomaly-scoring implementation with score threshold $= 1.8$. In our end-to-end wrapper used together with the practical-deployment efficiency analysis, we set remove count $= 1$, Top-$K = 5$, max clean hits = 50, max new tokens = 64, and judge max new tokens = 16.

\subsection*{Perturbation Types and Hyperparameter Settings}

\paragraph{Token dropout.}
In this setting, we randomly drop document tokens by modifying the attention mask with a dropout probability of 0.10. The perturbation is applied only to the document side; the query is left unchanged. The [CLS] token is always preserved, and the implementation further ensures that at least one additional document token remains active after masking. 

\paragraph{Encoder dropout.}
In this setting, we activate the encoder's native internal dropout by switching the encoder to training mode during penalty estimation. We do not add any extra manually tuned dropout module. For Contriever, this corresponds to the pretrained model's default configuration, with hidden dropout 0.10 and attention dropout 0.10. 

\paragraph{Mixed perturbation.}
In the mixed setting, we combine encoder dropout and token dropout. Specifically, we enable the encoder's native internal dropout and simultaneously apply token dropout with probability 0.10 on the document side only.

\end{document}